\documentclass[letterpaper]{article} 
\usepackage{aaai23}  
\usepackage{times}  
\usepackage{helvet}  
\usepackage{courier}  
\usepackage[hyphens]{url}  
\usepackage{graphicx} 
\usepackage{natbib}  
\usepackage{caption} 
\usepackage{algorithm}
\usepackage{algorithmic}

\usepackage{newfloat}
\usepackage{listings}

\usepackage{graphicx}
\usepackage{wrapfig}

\usepackage{colortbl}
\usepackage{xcolor}         
\usepackage{amsmath}
\usepackage{amsfonts}

\DeclareCaptionStyle{ruled}{labelfont=normalfont,labelsep=colon,strut=off} 
\floatstyle{ruled}
\newfloat{listing}{tb}{lst}{}
\floatname{listing}{Listing}
\title{Prototypical Fine-tuning: Towards Robust Performance Under Varying Data Sizes}
\author{
    Yiqiao Jin\textsuperscript{\rm 1}\thanks{Work done during an internship at Microsoft Research Asia.},
    Xiting Wang\textsuperscript{\rm 2}\footnote{Xiting Wang is the corresponding author.},
    Yaru Hao\textsuperscript{\rm 2},
    Yizhou Sun\textsuperscript{\rm 3},
    Xing Xie\textsuperscript{\rm 2}
}

\affiliations{
    \textsuperscript{\rm 1} Georgia Institute of Technology, 
    \textsuperscript{\rm 2} Microsoft Research Asia,
    \textsuperscript{\rm 3} University of California, Los Angeles \\
    yjin328@gatech.edu, ~
    \{xitwan, yaruhao, xing.xie\}@microsoft.com, ~
    yzsun@cs.ucla.edu
}

\usepackage{bibentry}

\definecolor{Gray}{gray}{0.9}
\definecolor{LightCyan}{rgb}{0.88,1,1}

\begin{document}

\maketitle

\begin{abstract}
In this paper, we move towards combining large parametric models with non-parametric prototypical networks. We propose prototypical fine-tuning, a novel prototypical framework for fine-tuning pretrained language models (LM), which automatically learns a bias to improve predictive performance for varying data sizes, especially low-resource settings. Our prototypical fine-tuning approach can automatically adjust the model capacity according to the number of data points and the model's inherent attributes. Moreover, we propose four principles for effective prototype fine-tuning towards the optimal solution. Experimental results across various datasets show that our work achieves significant performance improvements under various low-resource settings, as well as comparable and usually better performances in high-resource scenarios. 
\end{abstract}

\section{Introduction}

Pretrained language models (LM) have achieved substantial success on a variety of NLP applications~\cite{brown2020language}. 
Their high discriminative power is partly attributed to a weak inductive bias, which poses little constraints on the model expressivity, but leads to potential overfitting and local optimum, especially in low-resource settings~\cite{mccoy2019right, belinkov2020variational, feldman2020does}. 
On the other hand, non-parametric models, such as prototypical networks
~\cite{snell2017prototypical, yang2018robust, shao2017robust}, 
introduce strong inductive biases and explicitly model the inter-class and intra-class relations within the data, which is especially useful when training data is limited. 
These methods are promising for improving generalization and increasing sample efficiency~\cite{snell2017prototypical, allen2019infinite}. 
However, as strong inductive biases restrict the hypothesis set of function approximators, these models introduce false assumptions that may limit the expressivity of models as the number of data points increases~\cite{baxter2000model}. 

In this paper, we move towards combining large parametric models with non-parametric prototypical networks so that we automatically learn a bias to improve predictive performance on varying sizes of datasets, especially for low-resource settings. 
Achieving this goal is non-trivial due to two major challenges. 
First (\emph{adaptation}), how to introduce prototype modeling into large-scale language models to jointly ensure superior performances under low-resource scenarios while mitigating the over-reliance of prototype learning on strong inductive biases?
Second (\emph{capacity}), different datasets have diverse levels of complexity, thus requires different volumes of prototypes and parameters to properly represent the data distribution~\cite{mettes2019hyperspherical}. Choosing model capacity generally requires heuristics and knowledge about the data distribution~\cite{allen2019infinite}.
How to determine the number of prototypes automatically by jointly considering complexity of the data distribution and the model's inherent attributes to ensure robust performances on datasets with varying sizes?\looseness=-1

To solve these two challenges, we propose \textbf{\underline{P}rototypical \underline{Fi}ne-\underline{t}uning (PFit)}.
Our method represents the data within each class as a set of prototypes, each modeled as a mixture component, while learning the number of components automatically based on the model capacity and the complexity of data distribution. 
At the initial stage of training, our method is equivalent to ProtoNet~\cite{snell2017prototypical}, which ensures a high-quality initialization of the prototypes and facilitates convergence towards a globally optimal solution. 
As training proceeds, \emph{PFit} chooses from fitting each new example with existing prototypes or initiating a new prototype based on the data point, according to whether the example is in-distribution. 
We allow the number of prototypes to grow and shrink dynamically with respect to the data complexity. 
In all, our method ensures a hypothesis space large enough for viable solutions to the tasks being learned~\cite{baxter2000model}, but small enough for generalization.
Thus, our method is well-suited for both data-scarce and data-rich tasks, especially offering substantial performance gain under low-resource scenarios. The contributions of this paper are summarized as follows:\looseness=-1 

1) We propose \emph{Prototypical Fine-tuning (PFit)}, a novel prototypical method for fine-tuning pretrained language models by integrating the strengths of bayesian non-parametrics with parametric pretrained LMs, leading to superior predictive performance under varying data sizes.

2) We propose four principles for effectively leveraging non-parametric methods 
, which are mixture prototype initialization, infinite mixture prototype creation, adaptive prototype refinement, and dynamic diversity regularization, so that our method maintains a compact set of prototype embeddings conducive to its decision-making;

3) We conduct extensive experiments and show that \emph{PFit} achieves considerable performance improvements under varying data sizes, as well as comparable or better performances for high-resource settings. 
\section{Related Work}
\subsection{Prototypical Learning}
Prototypical Learning~\cite{mettes2019hyperspherical, snell2017prototypical, oreshkin2018tadam} 
is closely related to metric learning~\cite{goldberger2004neighbourhood, oh2016deep} and case-based reasoning~\cite{li2018deep, kim2014bayesian, kolodner1992introduction}.
It has been used for improving model interpretability~\cite{li2018deep, chen2019looks} and few-shot classification~\cite{snell2017prototypical, gao2019hybrid}. 
Most existing works combine bayesian non-parametrics with representation learning and learn a single prototype for each class~\cite{vinyals2016matching, snell2017prototypical}, based on the uni-modality assumption of embedded class examples. As an extension, multi-modal methods~\cite{hjort2010bayesian, rasmussen1999infinite, mensink2013distance, ackerman2014incremental} assume an infinite number of mixtures, but only a finite number are used to generate the data. In particular, infinite mixture prototype (IMP)~\cite{allen2019infinite, kulis2012revisiting} adaptively fits the data distribution through a varied number of clusters for each class. Notably, existing works lack an efficient mechanism to be dynamically scaled on datasets with varying sizes.

\subsection{Low-Resource}
Due to the cost of acquiring high-quality labeled data, researchers have been working on improving model performances under low-resource settings. A special case is few-shot learning~\cite{wang2021grad2task, tsimpoukelli2021multimodal, perez2021true} in which data is severely limited.
Earlier works address data sparsity through large unlabeled datasets~\cite{gunel2020supervised, artetxe2018unsupervised}, data augmentation
~\cite{salamon2017deep, fadaee2017data}, or the removal of redundant information in the pretraining representation~\cite{belinkov2020variational}. 
These methods usually introduce extra data or auxiliary models that increase the computational cost. 
Moreover, these works are usually designed for specific low-resource settings, but cannot robustly scale to larger datasets with performance guarantee~\cite{zaken2021bitfit}. 
In this work, we propose a non-parametric method suited for both low- and high- resource scenarios that achieves particularly superior performances under the former settings.
\section{Method}
In this section, we first introduce the general workflow for fine-tuning language models. Then, we present the proposed approach \emph{PFit} and show how it achieves superior performances under varying data sizes. 
\begin{figure*}[t]
\vspace{-3mm}
\center
\includegraphics[width=0.9 \textwidth]{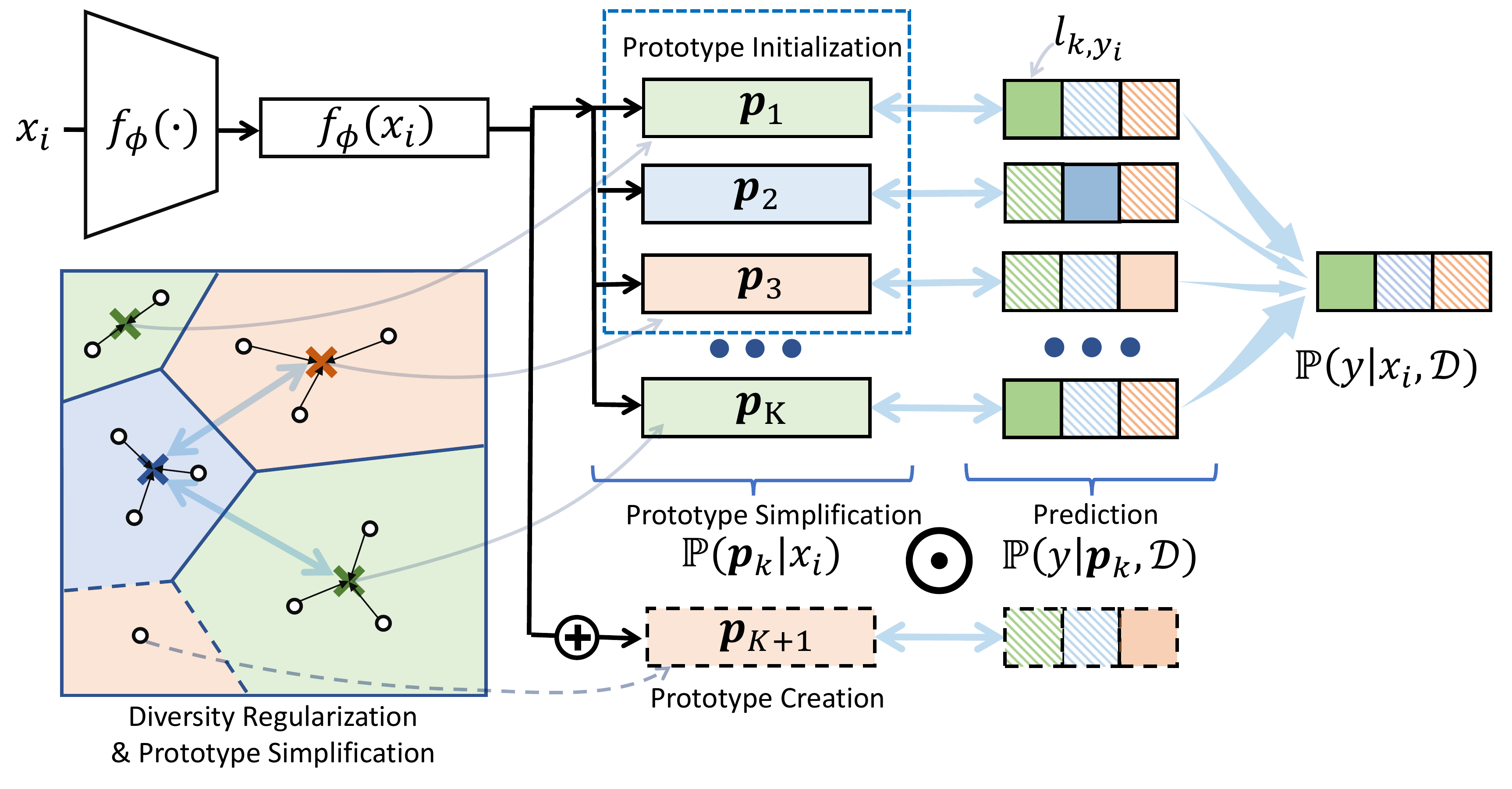}
\caption{Our proposed \emph{PFit} framework. During fine-tuning, the prediction of each example $x_i$ is given by individual predictions of prototypes weighted by their distance to $f_{\phi}(x_i)$. \emph{PFit} is able to dynamically add or prune prototypes, depending on whether existing prototypes can characterize the distribution of data.}
\label{fig:pfit}
\vspace{-3mm}
\end{figure*}
\subsection{Problem Statement}
\textbf{Input} The input for fine-tuning is a labeled dataset $\mathcal{D} = \{(x_{i}, y_{i})\}_{i=1}^N$, 
where each example $x_i \in X$ is an input text sequence, and $y_i \in \{1, \ldots, C\}$ is the ground-truth label. 

\noindent \textbf{Output} The outputs consist of 1) $\mathbb{P}(y \mid x_i)$, 
the prediction of each example $x_i$; 2) the fine-tuned model $f_{\phi}(\cdot): X \rightarrow \mathbb{R}^C$ that maps an input sequence to a probability distribution over $C$ classes. 

\subsection{Prototypical Fine-tuning}
Pretrained language models are fine-tuned in a fully parametric way~\cite{wu2022noisytune}. The inference is conducted in a case-by-case manner by computing $\mathbb{P}(y \mid x_i)$. 
This way, performances of LMs predominantly stem from their large-scale parameters, whereas they do not explicitly leverage the class distribution and instance-level similarity among training examples, which is especially useful when data is limited.

To alleviate this problem, we explicitly model the inter-class and intra-class relations by introducing \emph{prototypes}, which are embeddings in the same metric space as $f_{\phi}({x}_i)$ that abstracts the essential semantics of multiple $x_i$'s~\cite{snell2017prototypical}.
In our case, we learn a compact set of prototypes $\mathbf{p}_{1 \sim K}$ from $\mathcal{D}$ that serves as references for model prediction.
We split the inference into two components: 
1) \textbf{prototype prediction} of each $\mathbf{p}_k$ conditioned on the entire training set $\mathcal{D}$; 
2) \textbf{prototype importance} that measures the compatibility between each $\mathbf{p}_k$ and $x_i$.
\begin{equation}
\mathbb{P}(y \mid x_i, \mathcal{D})
= \sum_{k=1}^{K} \underbrace{\mathbb{P}(y \mid \mathbf{p}_k, \mathcal{D})}_{\substack{\text{Prototype}\\ \text{Prediction}}}
\underbrace{ \mathbb{P}(\mathbf{p}_k \mid x_i)}_{\substack{\text{Prototype}\\ \text{Importance}}}
\label{eq:pfit_joint_prob}
\end{equation}
When infering on a new unclassified exemplar, prototypical networks calculates joint prediction probability $\mathbb{P}(y \mid x_i, \mathcal{D})$ and efficiently adapts existing prototypes to the new data rather than deriving answers from scratch. Under this formulation, prototypical networks focus on critical aspects of the data distribution and filter irrelevant or noisy features that are detrimental to their decision-making~\cite{kolodner1992introduction} better at finding a globally optimal solution. 

\subsubsection{Prototypical Network}
Under the prototypical learning paradigm, each $x_i$ is encoded into a $D$-dimensional feature vector $f_{\phi}(x_i) \in \mathbb{R}^D$ and matched against a set of prototypes $\mathbf{p}_k$ in a learned latent space according to some distance metric $d(\cdot, \cdot)$.
The importance of prototype $\mathbf{p}_k$ with respect to $x_i$ is measured by: 
\begin{equation}
\mathbb{P}(\mathbf{p}_k \mid x_i) \propto \operatorname{exp}(-d(f_{\phi}(x_i), \mathbf{p}_k))
\end{equation}
where $d(\cdot, \cdot)$ is usually taken as the squared L2 distance. Prototypical methods~\cite{snell2017prototypical} calculate the prototype of each class $c$ as the averaged embedding of $\mathcal{D}_{c} \in \mathcal{D}$, or the support data within class $c$:
\begin{equation}
\mathbf{p}_{c}=\frac{1}{\left|\mathcal{D}_{c}\right|} \sum_{\left(x_{i}, y_{i}\right) \in \mathcal{D}_{c}} f_{\phi}(x_i)
\label{eq:protonet}
\end{equation}
Prediction is conducted by calculating the distance from $f_{\phi}(x_i)$ to each prototype vector.
\begin{equation}
\mathbb{P}(y=c \mid x_i)=\frac{\exp (-d(f_{\phi}(x_i), \mathbf{p}_{c}))}{\sum_{c^{\prime}} \exp (-d(f_{\phi}(x_i), \mathbf{p}_{c^{\prime}}))}
\label{eq:single_proto}
\end{equation}

\subsubsection{Our PFit Framework} 

Our goal is to learn a \textbf{\underline{P}rototypical \underline{Fi}ne-\underline{t}uning (PFit)} network with robust performances under both low- and high-resource settings. The training procedure is detailed in Alg.~\ref{alg:pfit}. 
Compared with a transformer-based model, PFit contains an additional prototype learning module $P$ that can flexibly adjust the capacity of its prototype embeddings $\mathbf{p}_k$ according to the data distribution. Our proposed PFit framework can be easily built on top of any backbone pretrained LMs. It pursues 4 approaches for robustness and expressivity under varying data sizes:

1) \emph{Mixture Prototype Initialization.} We leverage the high-level semantic and linguistic features in pretrained representations to initialize the prototypes for faster convergence;

2) \emph{Infinite Mixture Prototype Creation.} We extend infinite mixture prototype~\cite{allen2019infinite} to flexibly capture the complexity of the distribution. The prototype embeddings are fine-tuned along with the pretrained model in a data-driven manner;

3) \emph{Adaptive Prototype Simplification.} To improve generalization and efficiency, we leverage simple inductive biases and maintain a compact set of prototypes that sufficiently represent the data distribution. 

4) \emph{Dynamic Diversity Regularization.} 
We dynamically enforce diversity among prototypes to ensure the expressivity and efficiency of our prototypical framework.


\subsubsection{Mixture Prototype Initialization}
\label{sec:init}

At the start of training, a simple yet strong inductive bias can ensure robust generalization since the amount of data is relatively limited~\cite{snell2017prototypical}.
Previous studies have shown that transformer models encode high-level semantic and linguistic features in their representations~\cite{tenney2019bert, jawahar2019does, hao2020investigating, guan2019towards}, leading to superior performances in various downstream tasks~\cite{yang2022glue, jin2021towards, yang2022reinforcement}.
Therefore, we propose \emph{mixture prototype initialization} to leverage the rich semantics in the sentence representations~\cite{yang2022mutually}. Before fine-tuning starts, for each class $c$, we randomly sample a subset of $n$ examples with label $c$, and aggregate their $f_{\phi}(x_i)$ with $\phi$ frozen:
\begin{equation}
\mathbf{p}_c = \operatorname{Aggr}([f_{\phi}({x}_1), \ldots, f_{\phi}({x}_n)])
\end{equation}
In our case, we choose mean pooling as the aggregate function. We derive an initial set of $C$ prototypes for all classes.
Mixture prototype initialization is analogous to ProtoNet~\cite{snell2017prototypical} and assumes that the classes form uni-modal clusters. Such initialization can empower the prototypes with the semantics of transformer word representations with little computational cost.
Meanwhile, it does not require extra parameters or training, only the intrinsic features of language models.

\subsubsection{Infinite Mixture Prototype Creation}
\label{sec:prototype_creation}

As the training proceeds, existing prototype embeddings become less capable of assimilating new data points. 
Higher capacity is required to model the growing data complexity.
A straightforward remedy is to directly create a set of 
$K$ prototypes and automatically learn meaningful representations from the training data. However, such method is largely subject to the choice of model capacity, either underfitting or overfitting the data~\cite{snell2017prototypical, ren2018meta}. 
Second, in practice, this method usually yields highly similar prototypes~\cite{ming2019interpretable} which hampers the expressivity of the prototype module and leads to sub-optimal convergence.

Motivated by~\cite{allen2019infinite}, we seek informative samples from the training data that contribute to the diversity of the prototypes and the expressivity of the fine-tuned models. Let $P^c$ be the set of prototypes under class $c$. For each example $x_i \in \mathcal{D}_c$, if the minimum distance between $f_{\phi}(x_i)$ and any prototypes in $P^c$ exceeds a threshold $\lambda$, 
a new prototype is created based on $f_{\phi}(x_i)$. The threshold distance $\lambda$ is given by:
\begin{equation}
\lambda=2 \sigma \log \left(\frac{\alpha}{\left(1+\frac{\rho}{\sigma}\right)^{d / 2}}\right)
\label{eq:lambda}
\end{equation}
where $\sigma$ is the cluster variance that is learned jointly with $\phi$.
$\rho$ measures the standard deviation for the base distribution from which the cluster means are sampled.
$\alpha$ is a hyperparameter that controls the concentration of clusters in the Chinese Restaurant Process~\cite{wang2009variational}. 
This way, our approach can choose between fitting simple data distribution with low capacity and complex distribution with high capacity. 
As a further extension, we only allow the creation of prototypes after being fine-tuned for a certain number of steps. This way, the initial $C$ prototypes are sufficiently trained along with $\phi$ so that meaningful prototypes can be created. For computational and storage efficiency, we restrict the size of prototypical networks to $|P_{max}|$.

\subsubsection{Prediction} 
During fine-tuning, language models classify new examples by applying an MLP on the word representations. PFit instead directly learns a distribution over all $C$ classes for each $\mathbf{p}_k$.
During fine-tuning, PFit learns a distribution over all $C$ classes for each $\mathbf{p}_k$.
Specifically, each $\mathbf{p}_k$ directly correlates with a learnable vector $\mathbf{l}_k \in \mathbb{R}^C$, which gives a probability distribution over all $C$ classes:
\begin{equation}
\mathbb{P}(y = c \mid \mathbf{p}_k, \mathcal{D}) = \operatorname{softmax}_c(l_{kc});~~ \mathbf{l}_k = [l_{k, 1}, \ldots, l_{k, C}]
\label{eq:}
\end{equation}
During initialization, for each newly created prototype $\mathbf{p}_k$, $l_{k, y_i}$, the value in $\mathbf{l}_k$ that corresponds to the ground-truth class $y_i$, is initialized to $\beta$, a positive real number. The value for all other classes $l_{k, c}, c \in \{1, \ldots, C\}\backslash y_i$ are initialized to $-\beta$. During fine-tuning, $\mathbf{l}_k$ is optimized together with its corresponding $\mathbf{p}_k$ in a data-driven manner.
In order to keep the predicted label constant, $l_{k, y_i}$, which corresponds to the ground-truth class $y_i$, is constrained in $[0, +\infty)$. Predictions for other classes 
are constrained in $(-\infty, 0]$.
Compared with existing prototypical approaches which usually use a constant value for prototype predictions, our method can achieve superior expressivity.

The prototype importance $z_{i, k}$ is given by the normalized Gaussian density:
\begin{equation}
\mathbb{P}(\mathbf{p}_k \mid x_i) = z_{i, k}=\frac{\mathcal{N}(f_{\phi}(x_{i}); \mathbf{p}_{k}, \sigma_{k})}{\sum_{k'} \mathcal{N}(f_{\phi}(x_{i}) ; \mathbf{p}_{k'}, \sigma_{k'})}
\label{eq:pfit_inference}
\end{equation}
The joint probability of $\mathbf{p}_k$ predicting class $c$ is given by Eq.~\ref{eq:pfit_joint_prob}.
Intuitively, our model is analogous to a retrieval system or an attention-based model~\cite{vaswani2017attention}, in which a 
query example is mapped against a set of key-value pairs. 
The weight assigned to each value is computed by a compatibility function between the query and each key. 
The prediction result is a linear combination of values weighed by the compatibility between the query example $f_{\phi}(x_i)$ and existing prototypes $\mathbf{p}_k \in P$.

\begin{algorithm}[ht!]
\caption{Prototypical Fine-tuning}
\label{alg:pfit}
\textbf{Input}: $\mathcal{D} = (x_i, y_i)$, where each $y_i \in \{1, \ldots, C \}$

\textbf{Output}: prototypes $\mathbf{p}_{k}$ and the classification of each example $\mathbb{P}(y \mid \mathbf{p}_{1\sim K},  x_i)$
\begin{algorithmic}[1]
\STATE Perform \textbf{Mixture Prototype Initialization} 
\FOR {\textrm{minibatch} $B_r \in D$ }    
 { 
    \STATE {Perform \textbf{Infinite Mixture Prototype Creation} and} estimate $\lambda$ according to Eq.~\ref{eq:lambda} 
    \FOR {$x_{i} \in B_r$}
    \STATE{  
        \FOR{ $k \in \{1, \ldots, K\}$ }  
               \STATE{
                 \COMMENT{\texttt{Compute the distance from $x_{i}$ to all existing prototypes}}
                 \STATE Calculate $d_{i, k} = d(f_{\phi}(x_i), \mathbf{p}_{k})$ for $\mathbf{p}_{k} \in P^{y_{i}}$, and $d_{i, k} =  +\infty$ for $\mathbf{p}_{k} \notin P^{y_{i}}$
                 
               }
           \ENDFOR

            \IF { $\operatorname{min}_{k}d_{i,k} > \lambda$ }
             
                 \STATE { Create the $K+1$-th prototype $\mathbf{p}_{K+1}$ using $f_{\phi}(x_i)$; Increment $K$ by 1}
            \ENDIF
           
        }
    \ENDFOR
     \IF { $\operatorname{min}_{k}d_{i,k} > \lambda$ }
     
         \STATE { Create the $K+1$-th prototype $\mathbf{p}_{K+1}$ using $f_{\phi}(x_i)$; Increment $K$ by 1}
     \ENDIF
    
}
\ENDFOR
\end{algorithmic}
\end{algorithm}

\subsubsection{Adaptive Prototype Simplification}
\label{sec:pfit_simplification}
\label{sec:prototype_simplification}
Simple inductive biases usually imply robust generalization to varying datasets~\cite{baxter2000model}.
To improve generalization and efficiency, we propose \emph{adaptive prototype simplification} to ensure that a compact set of prototypes sufficiently represent the data distribution. 
As prototypes are created with examples that are sufficiently distinct from existing prototype embeddings, a minority of prototypes may be created by outliers that are rarely observed in the training data. These prototypes may fail to generalize to new data and are likely to introduce noisy features or spurious correlations. 
To determine unimportant prototypes, a simple way is to calculate the average prototype importance $z_{i, k}$ in Eq.~\ref{eq:pfit_inference} between each $\mathbf{p}_k$ and $f_{\phi}({x}_i)$ in the training set. 
This leads to two potential issues: 1) acquiring $f_{\phi}(x_i)$ and calculating $z_{i, k}$ among $f_{\phi}(x_i)$ and $\mathbf{p}_k$ (Eq.~\ref{eq:pfit_inference}) requires $O(N|P_{max}|)$, both of which are computationally expensive. 2) a direct remedy is to store all $z_{i, k}$ calculated during training and use them during simplification. However, the actual value of $z_{i, k}$ is constantly changing, as 
$f_{\phi}(x_i)$ and $\phi$ are continuously updating.

To solve the above two challenges, we introduce a sliding window of length $\delta$, which records all $z_{i, k}, i \in [T - \delta, T]$ at any training step $T$. 
Upon prototype simplification, we prune prototypes with an average prototype importance $\mathbb{P}(\mathbf{p}_k \mid x_i)$ that fall below a threshold $\varepsilon$:
\begin{equation}
\frac{1}{\delta} \sum_{i} \omega(i, T, \delta) z_{i, k} < \varepsilon
\end{equation}
$\varepsilon$ is a hyperparameter controlling the minimum average importance of $\mathbf{p}_k$ on all $f_{\phi}(x_i), i \in [T-\delta, T]$. 
$\omega(i, T, \delta) = \frac{1}{\delta}(i - T + \delta)$ is a linear discount factor that assigns higher importance to data points closer to $T$ in the sliding window. 
Our method not only avoids storing and indexing the entire set of representations, but also considers that, at any time step $T$, $f_{\phi}(x_i)$ that are computed closer to $T$ are more accurate. 
During each training epoch, prototype simplification is performed for a small number of $M$ times per epoch.

\subsubsection{Comparison to Previous Works} 
Previous works in few-shot learning~\cite{snell2017prototypical, allen2019infinite} use an alternative way to ensure the efficiency of prototypes. They compute a new prototype for each mixture component using the episode-end cluster mean, whereas in our model fine-tuning scenario, constant re-estimation of $\mathbf{p}_k$ will break the continuity of the model's learning, leading to poor convergence.
In contrast, our approach carefully considers the dynamically changing nature of example representation; global information, the discount factor $\omega(i, T, \delta)$ also mitigates overfitting a small batch.


\subsubsection{Optimization \& Extension} 
\textbf{Dynamic Diversity Regularization} 
We dynamically control the diversity among the prototype embeddings by restricting the minimum pairwise distance among $\mathbf{p}_k$'s to $\lambda$. 
As $P$ grows, $\lambda$ increases, enforcing higher pairwise difference among prototypes so that PFit maintains a set of meaningful prototypes out of $f_{\phi}(x_i)$ that sufficiently adds to the diversity of $P$ and avoids redundancy~\cite{wu2022adversarial} in the prototypes.
\begin{equation}
\mathcal{L}_{div}=\sum_{j=1}^{K} \sum_{k=j+1}^{K} \max \left(0, \lambda-\left\|\mathbf{p}_{j}-\mathbf{p}_{k}\right\|_{2}\right)^{2}
\end{equation}

\subsubsection{Extension to Regression} Our approach can be readily applied to regression tasks by partitioning the range of $y_i$ into a number of $N_{reg} \in \mathbb{N}^{*}$ bins and sample $n$ examples from each bin. Each prototype prediction $\mathbf{l}_{k}$ becomes a real number computed as the averaged ground-truth $y_i$ in each bin. Upon inference, we compute $\hat{y}_i$ as $\hat{y}_i = \sum_{k}z_{i, k} \mathbf{l}_{k}$.

\subsubsection{Optimization} We minimize cross-entropy loss for classification and mean squared error for regression. The full objective $\mathcal{L}$ linearly combines $\mathcal{L}_{0}$ and $\mathcal{L}_{div}$ using a hyperparameter $\rho_d$.
\begin{align*}
\mathcal{L}_{0} &= \begin{cases} \frac{1}{N}\sum_{i=1}^{N} \sum_{k=1}^K - y_{i,k} \log (\hat{y}_{i,k}) & \text{(classification)} \\
\frac{1}{N}\sum_{i=1}^{N}(y_i-\hat{y}_i)^{2} & \text{(regression)}
\end{cases} \\
\mathcal{L} &= \mathcal{L}_{0} + \rho_{d}\mathcal{L}_{div}
\label{eqn:loss_ce}
\end{align*}

\subsection{Interpretability}
\label{subsec:interpret}
There are 3 ways to interpret the prototype embeddings $\mathbf{p}_k$: 1) jointly train a decoder that transforms $\mathbf{p}_k$ back to the input space~\cite{li2018deep}; 2) project $\mathbf{p}_k$ onto the nearest sequence in the input space~\cite{ming2019interpretable}; 3) project $\mathbf{p}_k$ onto examples that are sufficiently close to the prototype within a threshold 
$\tau$: $\{x_i \mid  \| \mathbf{p}_{k} - f_{\phi}(x_{i})\|_2 < \tau  \}$.



\section{Experiment}

\textbf{Implementation Details} We apply a batch size of 32 and use $n=8$ examples to initialize each prototype. The sequence length is set to 128. We first conduct a grid-search on $\alpha$ within $\{1, 0.5, 0.1, 0.05, 0.01\}$ , $\rho_d$ in $\{1e-4, 1e-5, 1e-6\}$, and $\epsilon$ in $\{1e-3, 3e-3, 1e-2\}$ using a subset of SST-2 with 1000 data points, and fix $\alpha=0.1$, $\rho_d=1e-5$, and $\epsilon = 1e-3$ to conduct all experiments without further tuning.
We use the Adam optimizer~\cite{kingma2015adam} without weight decay. For both PFit and the baselines, we search the best learning rate from $[1e-5, 2e-5, 3e-5, 5e-5, 1e-4]$. Other hyperparameters of the baselines are set to their default values in their original papers. 
All experiments were run for a maximum of 5 epochs with early stopping. For hardware usage, experiments are run on servers with NVIDIA Tesla P100 (16GB GPU memory) or V100 (32GB GPU memory).

\subsection{Robustness Across Varying Data Sizes}
\label{sec:eval_robustness_data_sizes}
\subsubsection{Experiment Setup} Our goal is to demonstrate the superior performances of PFit across a wide range of data sizes, especially under low-resource scenarios.
Thus, we evaluate our model across a diverse spectrum of data sizes using 4 representative datasets from the General Language Understanding Evaluation (GLUE) benchmark~\cite{wang2019glue}, including QNLI, MNLI-m, and MNLI-mm, 3 textual entailment datasets, as well as SST-2, a common benchmark for sentiment analysis. The datasets are chosen following~\cite{gunel2020supervised} so that both single sentence and sentence-pair classification datasets are included.
We report our results on MNLI-m/mm in Tab.~\ref{tab:mnli} and SST-2/QNLI in Tab.~\ref{tab:sst2}. We validate the performances using a moderate-sized model BERT-Base~\cite{devlin2019bert} 
as our backbone model. We randomly sample 100 to 5000 examples from the training set as our training set, taking into account the class distribution of the full training set, and test our model on the full dev set. We conduct experiments with 5 different seeds and report the accuracy and standard deviation. 

\subsubsection{Baselines} We compare our PFit method against 6 baselines, including: 1) 4 non-parametric networks built on top of the backbone language models, including DP-Means~\cite{kulis2012revisiting}, ProtoNet~\cite{snell2017prototypical}, ProSeNet~\cite{ming2019interpretable}, and IMP~\cite{allen2019infinite}; 2) VIBERT, a network designed for varying-resource fine-tuning~\cite{belinkov2020variational}; 3) BSS, a regularization approach
to suppress untransferable spectral components and improve fine-tuning~\cite{wei2021pretrained} 4) the original LM.

\begin{table*}
\centering
\small
\begin{tabular}{r|lllllll|l|l} 
\hline
\hline
\multicolumn{1}{l|}{MNLI-m}  & DPMeans                         & ProtoNet                        & ProSeNet                        & IMP                               & BSS                             & ViBERT                          & Raw                             & PFit                            & Impr.
  (b) \% \\ 
\hline
100                          & 38.6 $\pm$ 2.1 & 35.5 $\pm$ 0.6 & 36.3 $\pm$ 1.2 & 36.7 $\pm$ 0.7   & 36.5 $\pm$ 0.9 & 39.3 $\pm$ 1.1 & 33.4 $\pm$ 0.8 & 40.6 $\pm$ 2.2 & 3.2          \\
200                          & 39.1 $\pm$ 1.3 & 37.5 $\pm$ 1.6 & 38.6 $\pm$ 0.4 & 41.2 $\pm$ 1.3   & 40.0 $\pm$ 0.7 & 40.1 $\pm$ 0.4 & 35.3 $\pm$ 0.9 & 42.4 $\pm$ 1.9 & 2.8          \\
500                          & 41.4 $\pm$ 1.5 & 43.2 $\pm$ 0.5 & 44.7 $\pm$ 0.8 & 47.5 $\pm$ 0.8   & 50.1 $\pm$ 0.5 & 49.2 $\pm$ 0.7 & 38.5 $\pm$ 0.8 & 52.9 $\pm$ 1.8 & 5.6          \\
1000                         & 53.0 $\pm$ 1.0 & 50.5 $\pm$ 0.3 & 52.2 $\pm$ 0.4 & 55.3
  $\pm$ 0.4 & 57.4 $\pm$ 0.4 & 57.1 $\pm$ 0.3 & 49.0 $\pm$ 0.7 & 58.5 $\pm$ 1.0 & 1.9          \\
2000                         & 62.5 $\pm$ 0.7 & 59.4 $\pm$ 0.7 & 59.0 $\pm$ 0.2 & 66.0 $\pm$ 0.3   & 65.6 $\pm$ 0.5 & 60.6 $\pm$ 0.3 & 59.6 $\pm$ 0.3 & 67.0 $\pm$ 0.7 & 1.5          \\
5000                         & 67.8 $\pm$ 0.5 & 67.4 $\pm$ 0.4 & 66.9 $\pm$ 0.1 & 71.1 $\pm$ 0.2   & 72.0 $\pm$ 0.1 & 71.4 $\pm$ 0.3 & 66.3 $\pm$ 0.1 & 72.3 $\pm$ 0.6 & 0.5          \\ 
\hline
\multicolumn{1}{l|}{MNLI-mm} &                                 &                                 &                                 &                                   &                                 &                                 &                                 &                                 &              \\ 
\hline
100                          & 38.8 $\pm$ 0.7 & 36.4 $\pm$ 0.4 & 36.8 $\pm$ 1.2 & 37.2 $\pm$ 1.1   & 35.2 $\pm$ 0.7 & 36.8 $\pm$ 1.8 & 34.1 $\pm$ 0.7 & 39.6 $\pm$ 2.1 & 1.8          \\
200                          & 39.5 $\pm$ 0.5 & 39.4 $\pm$ 1.2 & 38.6 $\pm$ 1.2 & 41.0 $\pm$ 0.2   & 41.5 $\pm$ 0.7 & 39.3 $\pm$ 0.8 & 35.7 $\pm$ 0.2 & 43.2 $\pm$ 1.8 & 4.0          \\
500                          & 46.4 $\pm$ 0.4 & 47.6 $\pm$ 1.1 & 40.7 $\pm$ 1.3 & 49.0 $\pm$ 2.2   & 49.5 $\pm$ 2.0 & 46.4 $\pm$ 1.4 & 45.9 $\pm$ 2.1 & 54.9 $\pm$ 1.0 & 11.1         \\
1000                         & 53.6 $\pm$ 0.6 & 55.4 $\pm$ 0.3 & 52.2 $\pm$ 0.4 & 54.8 $\pm$ 0.3   & 56.1 $\pm$ 0.2 & 48.6 $\pm$ 0.4 & 50.1 $\pm$ 0.3 & 61.5 $\pm$ 0.6 & 9.6          \\
2000                         & 64.3 $\pm$ 0.8 & 62.4 $\pm$ 0.3 & 61.0 $\pm$ 0.3 & 67.1 $\pm$ 0.2   & 64.5 $\pm$ 0.5 & 60.9 $\pm$ 0.3 & 58.6 $\pm$ 0.2 & 67.8 $\pm$ 0.2 & 1.1          \\
5000                         & 69.8 $\pm$ 0.3 & 70.0 $\pm$ 0.1 & 68.9 $\pm$ 0.2 & 71.5 $\pm$ 0.4   & 71.7 $\pm$ 0.3 & 68.3 $\pm$ 0.2 & 69.2 $\pm$ 0.3 & 71.9 $\pm$ 0.1 & 0.4          \\
\hline
\hline
\end{tabular}
\caption{Results on MNLI-m/mm with varying data sizes for BERT-Base. ``Raw'' indicates the original language model. The leftmost column shows the number of examples used in training. \textit{Impr.(b)} indicates the improvement over the best baseline}
\label{tab:mnli}
\end{table*}

\subsubsection{Model Performances}
The definition of low-resource settings varies across data sets~\cite{belinkov2020variational, gao2018low}. 
We follow existing literature~\cite{jin2020discrete, belinkov2020variational} to define low-resource settings as datasets with $\le$ 5000 data points.
Compared with the baselines, our PFit model consistently achieves the best performance among all datasets and data sizes.
For MNLI-m/mm, our method leads to a maximum of 5.6 and 11.1 percent improvement (Tab.~\ref{tab:mnli}) compared to the best baseline models. For QNLI/SST-2, our method leads to improvement of 5.1 and 3.8 percent (Tab.~\ref{tab:sst2}).

We also observe that the most significant improvement (Tab.~\ref{tab:sst2}) occurs with smaller datasets (100-200 examples) on SST-2, compared with medium-sized datasets (500-2000 examples) on QNLI and MNLI. This is potentially because sentiment analysis has a larger reliance on lower-level phrasal information and surface features, which are encoded in the pretrained word representations~\cite{jawahar2019does} and can be directly leveraged in the construction of prototypes. Our proposed mixture prototype initialization can effectively leverage the rich phrasal and sentiment clues and assist PFit towards achieving a globally optimal solution. 
For more complex tasks such as textual entailment, larger data sizes are needed for modeling higher-level complex semantic features and class relationships.
For small-scale datasets, models with simpler inductive biases like ProtoNet~\cite{snell2017prototypical} may achieve slightly worse but more stable performances with less parameters. On larger datasets, prototypical models with flexible capacity~\cite{allen2019infinite, kulis2012revisiting} generally outperform methods with fixed capacity~\cite{snell2017prototypical, ming2019interpretable}. 


\begin{figure}[t]
\vspace{-1mm}
\center
\includegraphics[width=0.21 \textwidth]{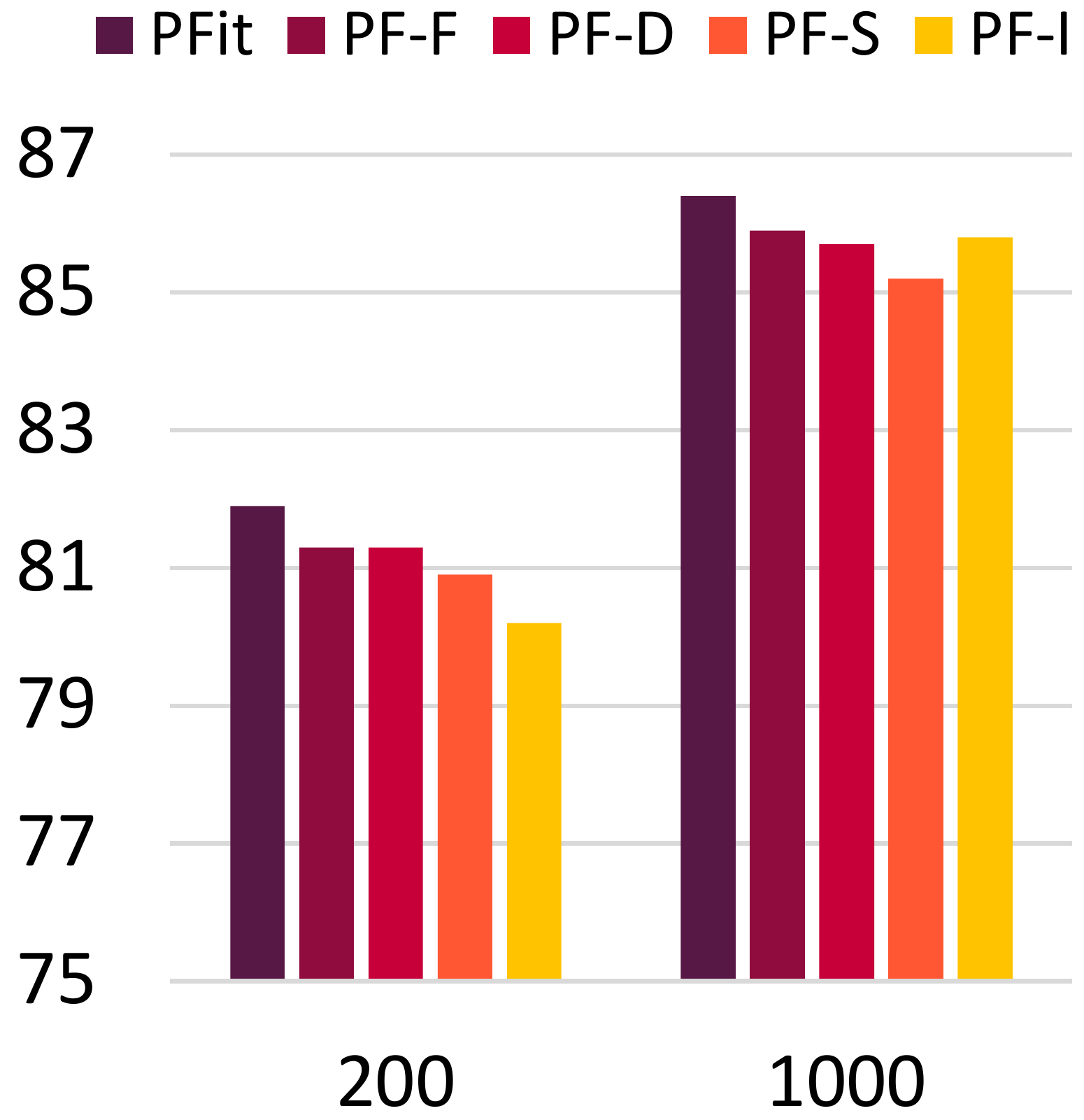}
\includegraphics[width=0.22 \textwidth]{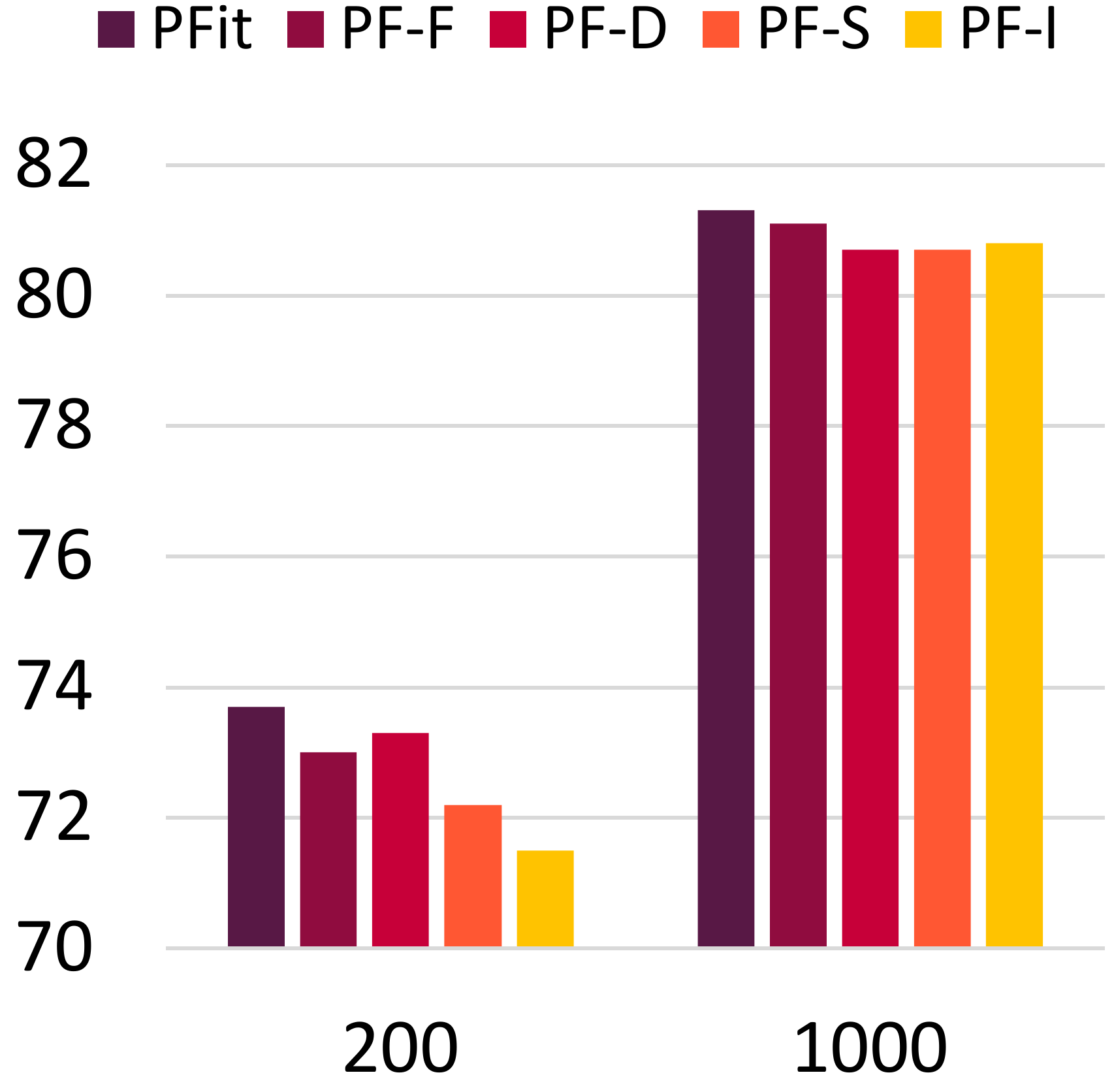}
\caption{Results of variations of \emph{PFit} model on SST-2 (Left) and QNLI (Right) with 1000 data points.}
\label{fig:ablation}
\vspace{-4mm}
\end{figure}

\begin{table}
\centering
\hspace{2mm}
\vspace{-1mm}
\begin{tabular}{l|lll}
\hline
\hline
Runtime & SST-2        & QNLI        & MNLI        \\ \hline
IMP     & 523         & 816         & 3139        \\
raw     & 18          & 28          & 105         \\
\hline
PFit    & 18          & 29          & 109         \\ \hline
\hline
\end{tabular}
\caption{Comparison of per-epoch runtime (in minutes) among the original model, IMP~\cite{allen2019infinite}, and PFit 
with batch size of 16.}
\label{tab:time}
\vspace{-4mm}
\end{table}


\begin{table*}
\centering
\small
\begin{tabular}{r|lllllll|l|l} 
\hline
\hline
\multicolumn{1}{l|}{SST-2} & DPMeans                         & ProtoNet                          & ProSeNet                          & IMP                               & BSS                               & ViBERT                            & Raw                               & PFit                              & Impr. (b) \% \\ 
\hline
100                        & 69.0 $\pm$ 0.7 & 74.9 $\pm$ 1.2   & 72.8 $\pm$ 0.9   & 74.7 $\pm$ 0.5   & 66.4 $\pm$ 0.7   & 67.5 $\pm$ 1.1   & 64.7 $\pm$ 0.9   & 78.7 $\pm$ 2.2   & 5.1        \\
200                        & 69.6 $\pm$ 2.5 & 72.9
  $\pm$ 0.6 & 75.5
  $\pm$ 1.2 & 80.0
  $\pm$ 1.3 & 68.6
  $\pm$ 0.4 & 68.7
  $\pm$ 0.6 & 67.1
  $\pm$ 1.2 & 81.9
  $\pm$ 1.1 & 2.4        \\
500                        & 73.9 $\pm$ 0.6 & 80.7
  $\pm$ 0.9 & 80.6
  $\pm$ 0.6 & 82.3
  $\pm$ 0.6 & 69.8
  $\pm$ 0.7 & 74.4
  $\pm$ 0.1 & 77.7
  $\pm$ 0.6 & 84.6
  $\pm$ 0.9 & 2.8        \\
1000                       & 84.6 $\pm$ 0.6 & 84.8$\pm$
  0.4  & 85.2
  $\pm$ 0.6 & 84.8
  $\pm$ 0.7 & 84.3
  $\pm$ 0.6 & 85.6
  $\pm$ 0.3 & 86.0
  $\pm$ 0.3 & 86.4
  $\pm$ 0.6 & 1.0       \\
2000                       & 87.2 $\pm$ 0.3 & 85.2
  $\pm$ 0.4 & 86.8
  $\pm$ 0.5 & 87.5
  $\pm$ 0.4 & 87.9
  $\pm$ 0.4 & 87.4
  $\pm$ 0.1 & 87.9
  $\pm$ 0.6 & 88.5
  $\pm$ 0.2 & 0.6        \\
5000                       & 88.3 $\pm$ 0.5 & 87.0
  $\pm$ 0.3 & 88.2
  $\pm$ 0.1 & 89.6
  $\pm$ 0.2 & 89.4
  $\pm$ 0.1 & 89.1
  $\pm$ 0.3 & 89.6
  $\pm$ 0.2 & 90.7
  $\pm$ 0.2 & 1.2        \\ 
\hline
\multicolumn{1}{l|}{QNLI}  & ~                               & ~                                 & ~                                 & ~                                 & ~                                 & ~                                 & ~                                 & ~                                 & ~          \\ 
\hline
100                        & 56.0 $\pm$ 2.7 & 64.8 $\pm$ 0.4   & 63.2 $\pm$ 1.7   & 61.2 $\pm$ 2.0   & 59.0 $\pm$ 1.3   & 60.3 $\pm$ 1.2   & 56.4 $\pm$ 1.4   & 67.0 $\pm$ 2.4   & 3.4        \\
200                        & 67.2 $\pm$ 0.4 & 67.4
  $\pm$ 1.3 & 69.0
  $\pm$ 1.5 & 71.0
  $\pm$ 1.2 & 69.2
  $\pm$ 0.5 & 69.2
  $\pm$ 1.3 & 62.1
  $\pm$ 1.6 & 73.7
  $\pm$ 1.0 & 3.7        \\
500                        & 71.1 $\pm$ 0.2 & 76.6
  $\pm$ 0.5 & 77.0
  $\pm$ 1.1 & 77.0
  $\pm$ 0.6 & 75.8
  $\pm$ 0.5 & 72.7
  $\pm$ 0.1 & 68.4
  $\pm$ 0.2 & 78.1
  $\pm$ 0.3 & 1.3        \\
1000                       & 78.6 $\pm$ 0.1 & 80.2
  $\pm$ 0.4 & 78.3
  $\pm$ 0.5 & 78.0
  $\pm$ 0.2 & 77.6
  $\pm$ 0.3 & 75.9
  $\pm$ 0.2 & 73.3
  $\pm$ 0.3 & 81.3
  $\pm$ 0.2 & 3.8        \\
2000                       & 81.6 $\pm$ 0.2 & 80.7
  $\pm$ 0.1 & 81.4
  $\pm$ 0.1 & 81.3
  $\pm$ 0.2 & 82.1
  $\pm$ 0.2 & 80.9
  $\pm$ 0.2 & 77.6
  $\pm$ 0.1 & 82.6
  $\pm$ 0.3 & 0.6        \\
5000                       & 82.7 $\pm$ 0.1 & 82.9
  $\pm$ 0.2 & 84.3
  $\pm$ 0.1 & 84.3
  $\pm$ 0.3 & 84.5
  $\pm$ 0.1 & 84.4
  $\pm$ 0.2 & 82.5
  $\pm$ 0.1 & 84.9
  $\pm$ 0.3 & 0.4        \\
\hline
\hline
\end{tabular}
\vspace{-1mm}
\caption{Results on QNLI/SST-2 with varying data sizes for BERT-Base.}
\label{tab:sst2}
\vspace{1mm}
\centering
\begin{tabular}{|l|l|l|} 
\hline
Prototype A                                                                                                                      & Prototype B                                                                                                                                                                     & Prototype C    \\ 
\hline
\begin{tabular}[c]{@{}l@{}}I had to look away - this \\was god~\textbf{awful}.\end{tabular}                                      & It's of the quality of a lesser harrison ford movie                                                                                                                             & Bad.           \\ 
\hline
\begin{tabular}[c]{@{}l@{}}I~\textbf{have to say}~the star \\and director are the~\\\textbf{big problems}~here.\end{tabular}     & \begin{tabular}[c]{@{}l@{}}I can take infantile humor ...~\textbf{but}~this is the sort\\of infantile that makes you wonder about \\changing the director and ...~\end{tabular} & Very bad.      \\ 
\hline
\begin{tabular}[c]{@{}l@{}}Eventually, every idea in \\this film is flushed down \\the~\textbf{latrine}~of heroism.\end{tabular} & \begin{tabular}[c]{@{}l@{}}\textbf{I am sorry that}~i was unable to get the full \\brunt of the comedy .\end{tabular}                                                           & Yes, dull ...  \\
\hline
\end{tabular}
\caption{The top 3 closest examples of each learned prototype on SST-2. The closest examples all have ground-truth labels of ``negative.'' However, each prototype represents a group of semantically similar negative samples.}
\label{tab:prototype_explainability}
\end{table*}

\subsection{Complexity \& Runtime}
The time complexity for PFit is $O(N|P_{max}|+NM)$. This consists of $O(N|P_{max}|)$ for prototype creation and $O(NM)$ for prototype simplification. While prototypical methods like IMP can achieve satisfying performances, they may incur expensive computational costs.
IMP takes $O(N^2|P_{max}|)$ due to recomputing a collection of prototype vectors at the end of each training step. 
For most data sizes, IMP~\cite{allen2019infinite} outperforms other baselines and sometimes achieves comparable results as PFit. However, IMP is computationally inefficient for a feature-learning setting as it requires constant re-estimation of the prototypes. As shown in Tab.~\ref{tab:time}, fine-tuning IMP requires 523 and 816 minutes per epoch on SST-2 and QNLI, respectively. In contrast, PFit takes 18 and 29 minutes. Compared with the raw model, PFit achieves significant improvement over the original language model with only a marginal runtime increase.

\subsection{Ablation Study}
To demonstrate the effectiveness of each major component, we evaluate 4 variations of PFit:
1) PF-I removes mixture prototype initialization and randomly initializes $\mathbf{p}_k$;
2) PF-S disables \emph{adaptive prototype simplification};
3) PF-D applies no diversity regularization $\mathcal{L}_{div}$;
4) PF-F uses a fixed threshold for $\mathcal{L}_{div}$.
We observe that mixture prototype initialization benefits the training under low-resource settings. The performance drops most significantly when we disabled \emph{adaptive prototype simplification}. We conjecture that some prototypes were created from a minority of out-of-distribution samples, which may overfit older training data but fail to generalize to new data. 


\begin{figure}[t]
\vspace{-4mm}
\center
\includegraphics[width=0.4 \textwidth]{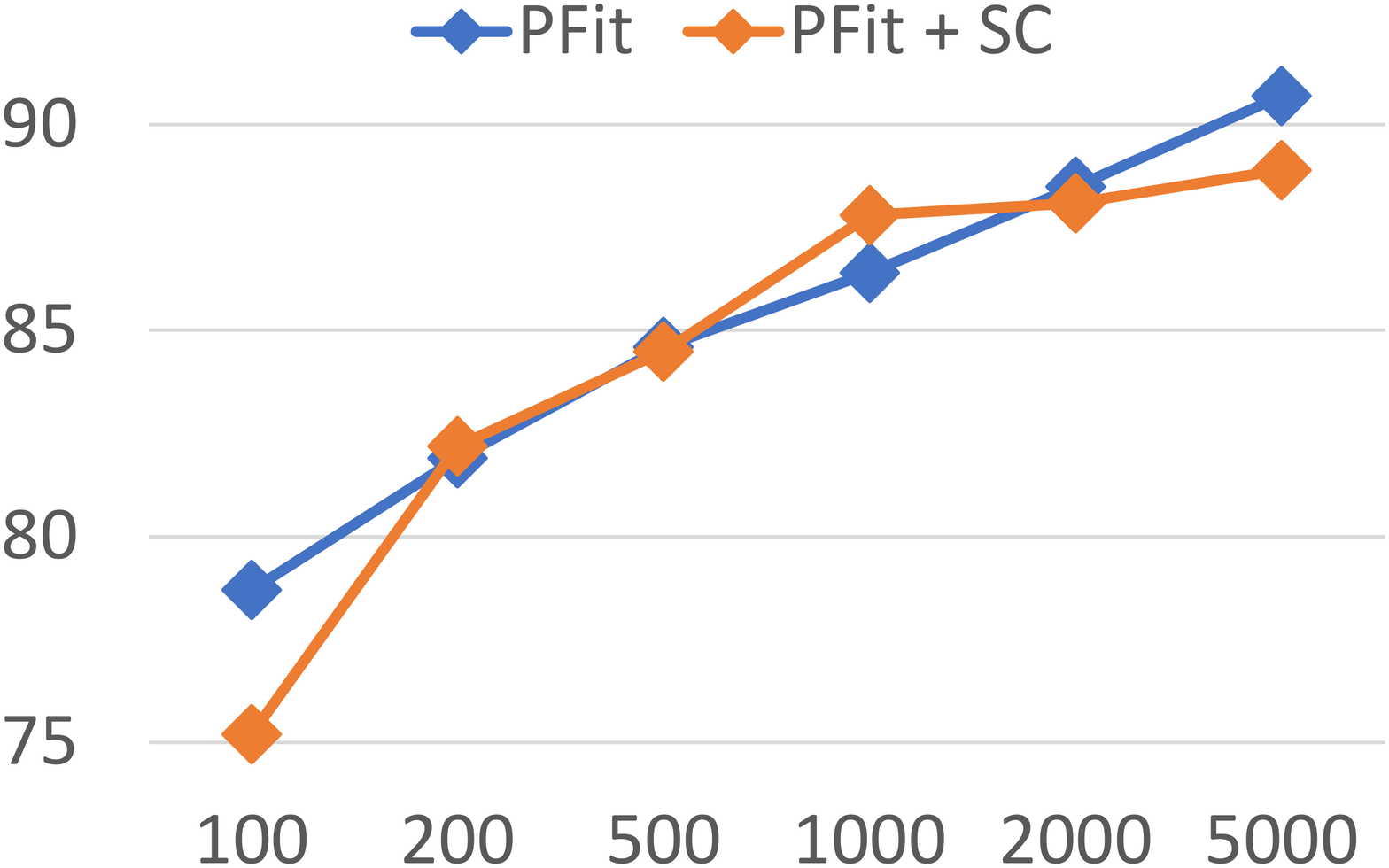}
\caption{Results for the original \emph{PFit} model and \emph{PFit} with skip connection (PFit-SC).}
\label{fig:pfit_skip_connection}
\vspace{-4mm}
\end{figure}

\subsection{Interpreting the Prototypes}

Besides performance improvements, prototypes can bring the extra advantage of explainability~\cite{lee2022self, gao2021learning, chen2021towards}. To interpret the prototype vectors, we project each prototype to its closest 10 examples. We observe that 1) prototypes manifest inter-class distinction as the classes are discriminable from the prototypes; and 2) prototypes demonstrate intra-class distinction~\cite{jin2022code} as there are clustering emergent within the prototypes.

\subsubsection{Inter-class Distinctions}
If we project each prototype onto its nearest examples, each prototype shows strong polarity towards one specific class. Here, for simplicity, we refer to prototypes that predict positive class as ``positive prototypes'' and those predicting negative as ``negative prototypes''. On SST-2, the average percentage of positive examples is 91\% for positive prototypes, and that of negative examples for negative prototypes is 87\%.

\subsubsection{Intra-class Distinctions}

Tab.~\ref{tab:prototype_explainability} is an example for the binary sentiment classification dataset SST-2. The top 3 closest examples to each prototype are shown. The closest examples all have ground-truth labels of ``negative'' However, each prototype represents a group of negative samples that are semantically similar. Examples of prototype A generally express stronger emotions (``awful'', ``big problem'', ``latrine''). Examples of B express weaker emotions. Examples of C mainly focus on succinct negative expressions. Methods such as ProtoNet~\cite{snell2017prototypical} only use one prototype for each class, which may not sufficiently characterize such intra-class distinctions.

We experimented with connecting non-contextualized BERT~\cite{devlin2019bert} embeddings with contextualized representations using a residual network, leading to performance improvements on SST-2 with 200 and 1000 examples, and decreases in other settings. We believe that merging the contextualized embeddings with non-contextualized signals without sacrificing performances can potentially increase the explainability of our prototypical framework. We will explore this method comprehensively in future works.

\section{Conclusion}
We propose Prototypical Fine-tuning (PFit), a prototypical framework for effectively fine-tuning pretrained language models. 
Extensive experiments across various datasets show that our work offers significant performance improvements for language models under varying data sizes, especially low resource settings.

\bibliography{aaai23}

\end{document}